%% file: iclr2022_conference.tex
\documentclass{article} 
\usepackage{iclr2022_conference,times}

\input{math_commands.tex}

\usepackage{algorithm}
\usepackage{algpseudocode}
\usepackage{overpic}
\usepackage{hyperref}
\usepackage{url}
\usepackage{booktabs}       
\usepackage{amsfonts}       
\usepackage{nicefrac}       
\usepackage{microtype}      
\usepackage{xcolor}         
\usepackage{graphicx}
\usepackage{bm}
\usepackage{amsmath}
\usepackage{amssymb}
\usepackage{todonotes}
\usepackage{subcaption}
\usepackage{multirow}
\usepackage{pifont}
\usepackage{overpic}

\newcommand{\name}{FABRIC}
\newcommand{\githublink}{https://github.com/sd-fabric/fabric}
\newcommand*\red{\color{red}}
\newcommand{\cmark}{\ding{51}}
\newcommand{\xmark}{\ding{55}}
\newcommand{\norm}[1]{\left\lVert#1\right\rVert}

\title{FABRIC: Personalizing Diffusion Models with Iterative Feedback}

\author{
Dimitri von Rütte\textsuperscript{1}\thanks{Equal contribution} \And Elisabetta Fedele\textsuperscript{1}\footnotemark[1] \And Jonathan Thomm\textsuperscript{1}\footnotemark[1] \And Lukas Wolf\textsuperscript{1}\footnotemark[1] \And
\vspace{-0.8cm}\\
\textsuperscript{1}ETH Zurich, Zurich, Switzerland \\
\texttt{\{dvruette, efedele, jthomm, wolflu\}@ethz.ch}
}

\iclrfinalcopy 

\begin{document}

\maketitle

\vspace{-0.8cm}
\begingroup
\setlength{\tabcolsep}{2pt}
\begin{figure}[h!]
    \centering
    \begin{overpic}[width=1\textwidth]{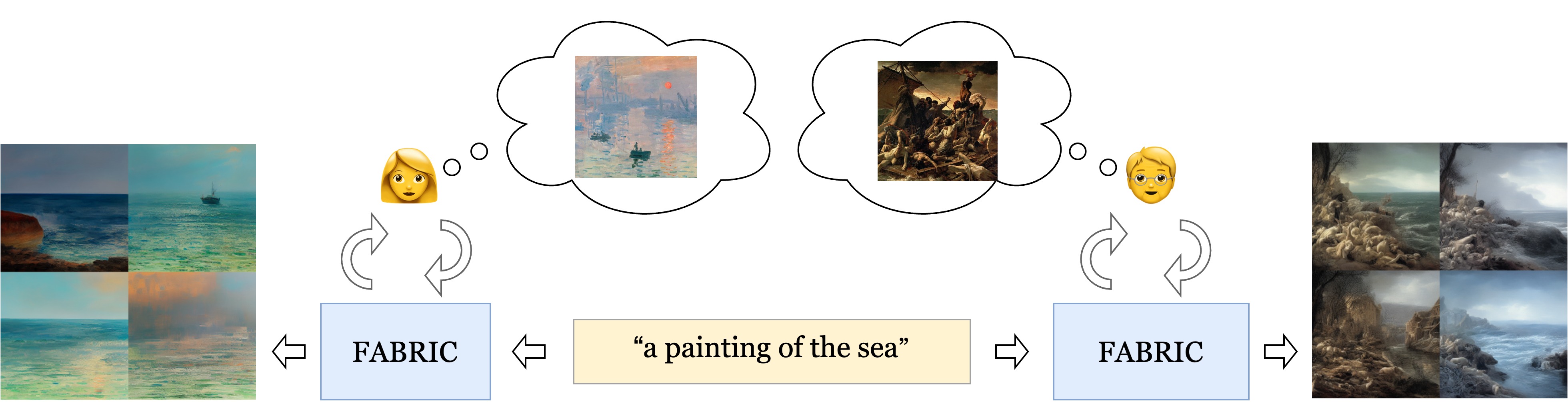}
    \put(-0.7,-2.1){{ \small \colorbox{white}{Output for user 1}}}
    \put(82.6,-2.1){{ \small \colorbox{white}{Output for user 2}}}
    \put(22.0,17.5){{ \small \colorbox{white}{user 1}}}
    \put(69.5,17.5){{ \small \colorbox{white}{user 2}}}
    \end{overpic}
    \vspace{0.1cm}\\
    \caption{Illustration of the proposed approach. FABRIC generates images based not only on text prompt, but also on user preferences expressed during multiple rounds of generation.}
\end{figure}
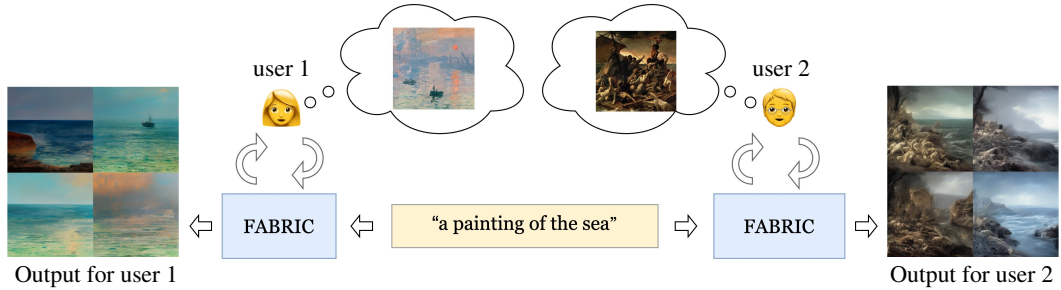
\endgroup
\begin{abstract}
    In an era where visual content generation is increasingly driven by machine learning, the integration of human feedback into generative models presents significant opportunities for enhancing user experience and output quality.
    This study explores strategies for incorporating iterative human feedback into the generative process of diffusion-based text-to-image models.
    We propose FABRIC, a training-free approach applicable to a wide range of popular diffusion models, which exploits the self-attention layer present in the most widely used architectures to condition the diffusion process on a set of feedback images.
    To ensure a rigorous assessment of our approach, we introduce a comprehensive evaluation methodology, offering a robust mechanism to quantify the performance of generative visual models that integrate human feedback. 
    We show that generation results improve over multiple rounds of iterative feedback through exhaustive analysis, implicitly optimizing arbitrary user preferences.
    The potential applications of these findings extend to fields such as personalized content creation and customization.
\end{abstract}

\section{Introduction}

\label{sec - intro}

The field of artificial intelligence (AI) has witnessed a surge in interest in generative visual models, primarily due to their transformative potential across a myriad of applications, encompassing content creation, customization, data augmentation, and virtual reality. These models leverage advanced deep learning methodologies, such as GAN \citep{goodfellow2014generative} and VAE \citep{kingma2022autoencoding}, to generate high-fidelity and visually compelling images from given inputs or descriptions \cite{brock2019large, razavi2019generating}. The significant advancements in generative visual models have catalyzed the exploration of novel possibilities in the realms of computer vision, natural language processing, and human-computer interaction \cite{radford2016unsupervised}.

Diffusion models, in particular, have emerged as a powerful tool in the field of image synthesis, often delivering results that are comparable to, or even exceed, those produced by GANs and VAEs \cite{ho2020denoising, rombach2022highresolution}. These models are characterized by their ability to generate a diverse array of visually coherent images, while demonstrating superior stability and reduced mode collapse during the training phase \cite{song2021scorebased}. This has led to their widespread adoption among researchers investigating the frontiers of generative visual modeling. Moreover, the utility of diffusion models extends beyond image synthesis, finding applications in various other domains such as inpainting, super-resolution, and style transfer \cite{chan2020glean, park2019semantic}.

Text conditioning serves as a crucial component of generative visual models, enabling them to synthesize images based on human-readable descriptions \cite{reed2016generative}. However, despite the robust capabilities of diffusion models in generating a wide spectrum of images, steering the model towards a specific desired output can pose challenges \cite{luccioni2023stable}. Users often embark on an iterative process of prompt refinement to achieve their intended results, and articulating personal preferences in the form of text can be a complex task. Nevertheless, users possess the ability to readily evaluate the quality of the generated images. This opens up the possibility of integrating sparse human feedback into the generative process, which can be harnessed to enhance the results and better align them with user preferences. Given the stability and controllability of the generative process offered by diffusion models, they present an ideal platform for the incorporation of human feedback to refine the text-to-image generation process.

In this work, we focus on this iterative workflow and propose a technique based on sparse feedback that aims to aid in steering the generative process towards desirable and away from undesirable outcomes.
This is achieved by using positive and negative feedback images (e.g. gathered on previous generations) to manipulate future results through reference image-conditioning.
Simply repeating the setup allows for iterative refinement of the generated images based on an arbitrary feedback source (including human feedback).
Our contributions are three-fold:
\begin{itemize}
    \item We introduce \name{} (\textbf{F}eedback via \textbf{A}ttention-\textbf{B}ased \textbf{R}eference \textbf{I}mage \textbf{C}onditioning) \footnote{The code is publicly available: \url{https://github.com/sd-fabric/fabric.git}}, a novel approach that enables the integration of iterative feedback into the generative process without requiring explicit training that can be combined with many other extensions to Stable Diffusion.
    \item We propose two experimental settings that facilitate the automatic evaluation of generative visual models over multiple rounds.
    \item Using these proposed settings, we evaluate FABRIC and demonstrate its superiority over baseline methods.
\end{itemize}

\begin{figure}
    \centering
    \includegraphics[height=4cm]{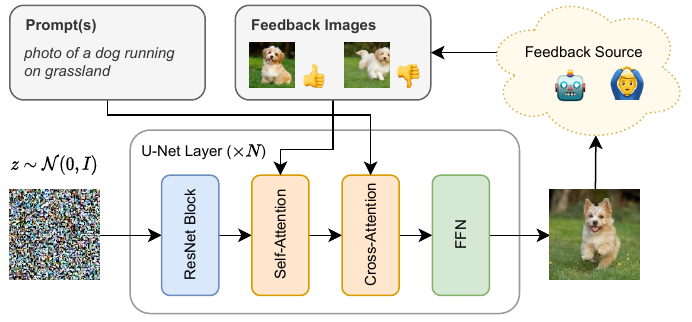}
    \caption{Illustration of the proposed approach. FABRIC improves generated results by incorporating user feedback through an attention-based conditioning mechanism.}
    \label{fig:fabric}
\end{figure}


\section{Related Work}
\label{sec - related work}

\subsection{Textual Inversion and Style Transfer}


A popular method for personalizing text-to-image diffusion models is textual inversion \citep{gal2023an, ruiz2023dreambooth}, a technique for learning semantic text embeddings from images depicting a common subject or style.
This enables the synthesis of photorealistic images in various scenes and conditions while preserving some set of desirable features, but requires multiple images that incorporate those features as well as additional training to learn the semantic embedding. \\\cite{mokady2022nulltext} introduce an accurate inversion technique for text-guided diffusion models, enabling text-based real-image editing capabilities. The proposed approach consists of two novel components: pivotal inversion for diffusion models and null-text optimization, allowing for high-fidelity editing of real images without tuning the model's weights. StyleDrop (\cite{sohn2023styledrop})is a novel method developed by Google Research for synthesizing images that adhere to a specific style using a text-to-image model. The method captures intricate details of a user-provided style, including color schemes, shading, design patterns, and local and global effects, and enhances the quality through iterative training with either human or automated feedback, outperforming other methods for style-tuning text-to-image models. \cite{xu2023promptfree} introduces a novel approach to incorporate user-provided reference images instead of text prompts. This approach, called Prompt-Free Diffusion, leverages a Semantic Context Encoder (SeeCoder) to transform these inputs into meaningful visual embeddings, which are then used as conditional inputs for a Text-to-Image (T2I) model, generating high-quality, customized outputs.

\subsection{Human Preference Modelling}

Recently, modeling human preferences in generative models experienced increased attention. In the domain of diffusion models, \citet{hps} provide a novel dataset containing almost 100 thousand images collected from over 25 thousand prompts. A sample from the dataset contains between two and four images and a particular user's choice when being offered this batch of images. The authors fine-tune a CLIP (\cite{radford2021learning}) model in order to classify chosen images against non-chosen ones. The \textit{Human Preference Score} (HPS) is derived from this classifier. In addition to that, the authors fine-tune a low-rank adaptation (LoRA, \cite{hu2021lora}) of stable diffusion (\cite{rombach2022highresolution}) that significantly improves the image quality.

\cite{kirstain2023pickapic} introduces Pick-a-Pic, a large-scale dataset of user preferences for text-to-image generation collected from real users of their web interface. The web app presents users with two generated images conditioned on their prompt and asks them to select their preferred option or indicate a tie if they have no strong preference. The rejected (non-preferred) image is then replaced with a newly generated image, and the process repeats. The dataset, containing 14,000 preferences, is used to train a scoring function called PickScore, which outperforms other available scoring functions in predicting user preferences, even surpassing expert human annotators. The authors advocate for the use of Pick-a-Pic prompts in evaluating text-to-image models, highlighting the dataset's potential for various applications such as model evaluation, image quality improvement, and text-to-image model enhancement.

\cite{fan2023dpok} proposes an approach called DPOK for fine-tuning text-to-image diffusion models using online reinforcement learning (RL). DPOK integrates policy optimization with KL regularization and updates pre-trained models using policy gradient to maximize feedback-trained reward. The authors demonstrate that DPOK generally outperforms supervised fine-tuning in terms of both image-text alignment and image quality.

\subsection{Iterative Feedback}


In the generative text-to-image workflow, the user typically thinks of a prompt, generates images with that prompt, inspects the results, makes adjustments to the prompt, and repeats this process until they're happy with the result. 
In order to remove this burden from the user, \cite{tang2023zerothorder} propose a novel zeroth-order optimization algorithm with theoretical guarantees for problems with black box objective functions evaluated through ranking oracles, such as Reinforcement Learning with Human Feedback (RLHF). 
The algorithm's effectiveness is demonstrated in improving image quality generated by a diffusion generative model using human ranking feedback, offering a promising alternative to existing RLHF methods.

To summarize, the literature of the field has predominantly focused on concept learning, style transfer from reference images, and modeling human preferences to fine-tune models accordingly. However, the iterative incorporation of human feedback into the model training process, a potentially significant aspect for enhancing model performance and alignment with human preference, remains relatively unexplored.

\section{FABRIC}
\label{sec:method}

We will now introduce our proposed method, FABRIC, which tackles the problem of incorporating multiple rounds of positive and negative human feedback into the generative process.

\paragraph{Attention-based Reference Image Conditioning}
FABRIC takes inspiration from a technique implemented in an update to the widely used ControlNet repository by \citet{zhang2023major} that introduces the ability to generate synthetic images similar to some reference.
This method exploits the self-attention module in the U-Net. The intuition is that the self-attention module's weights have learned to "pay attention" to other pixels in the image; therefore, adding additional keys and values from a reference image offers a way to inject additional information. 
In order to be compatible with the generated image at a specific time step of the denoising process, the reference latent image is partially noised up to the current timestep $t$ (using the standard random forward noising process:

\begin{equation}
    z_\mathrm{ref} \gets \sqrt{\bar{\alpha}_t} \cdot x_\mathrm{ref} + \sqrt{1 - \bar{\alpha}_t} \cdot \mathcal{N}(0, I)) 
\end{equation}

However, how to compute keys and values for a reference image in the U-Net layers is not immediately clear since simply concatenating the images does not work. Therefore, the keys and values for injection (or equivalently the hidden states right before, see Algorithm \ref{algorithm - fabric}) are computed by passing the noised reference image through the U-Net of Stable Diffusion. All keys and values are stored in the self-attention layers of the U-Net.

Then, for a particular U-Net denoising step of the user prompt and the current partially denoised latent image $z_{t+1}$, the stored keys and values are appended in the self-attention in the respective U-Net layers. This way, the denoising process can attend to the reference image and include the semantic information. By reweighting the attention scores (see Equation~\ref{eq:weightedattention}), we additionally obtain control over the strength of the reference influence.

Note that precomputing the hidden states of reference images requires an additional U-Net forward pass, roughly doubling the inference time.
Further, concatenating additional keys and values in the self-attention layer also increases inference time and results in scaling of the required memory that is quadratic in the number of feedback images (or linear if a memory-efficient implementation of attention is used) \citep{dao2022flashattention, rabe2022selfattention}.

\paragraph{Incorporating Feedback in the Generation Process}

The method outlined in the previous section incorporated a reference image into the generative process. We will now provide an extension to incorporate multi-round positive and negative feedback: 
Given a set of liked and disliked images, we do a separate U-Net pass for every liked and disliked image. The keys and values of the liked and disliked images are concatenated to the conditional and unconditional U-Net pass respectively (see classifier-free guidance, \cite{ho2022classifierfree}). Further, we reweight the attention scores with a factor depending on whether it is the conditional or the unconditional pass and the current time step in the denoising process. 

Unlike the ControlNet \citep{zhang2023major} approach, we also explore this linear interpolation which allows us to emphasize coarse features (having a large reference attention weight for $t\geq \alpha T$) or fine details from the reference (i.e., larger attention weight when $t\leq \alpha T$).
The feedback process can be scheduled according to the denoising steps. This allows for only including feedback in some of the denoising steps. We find that the best results can be achieved when incorporating the feedback in the first half of the denoising process. Additional experiments with adapted feedback schedules can be found in Section \ref{sec - adapting the feedback schedule}.

\paragraph{Extending to Iterative Feedback}
Now that we can incorporate both positive and negative feedback reference images, we adapt the algorithm for multiple rounds. Initially, we generate images with no feedback images, resulting in a vanilla Stable Diffusion generation. From those images, we append any set of \emph{liked} and \emph{disliked} images to the positive and negative feedback respectively. How this feedback is obtained is arbitrary, although we envision either a human or an automated process being the feedback source. Then, we generate a new batch of images using the given feedback. We repeat this process every round, extending the sets of liked and disliked images and refining the next batch according to them.

\section{Evaluation}

\subsection{FABRIC Models}

We conduct an evaluation of two versions of \name{} in our study. The first version, referred to as \name{}, utilizes the methodology outlined in Section~\ref{sec:method}, built upon a fine-tuned Stable Diffusion 1.5 checkpoint (dreamlike-photoreal-2.0).
The second version, named \name{}+HPS LoRA, further enhances the \name{} approach by applying it on top of the Low-Rank Adaptation (LoRA) of Stable Diffusion 1.5 based on Human Preference Score proposed by \citet{hps}. We opted to include the \name{}+HPS LoRA version in our evaluation due to its proven ability to generate images that closely match human preferences.

\subsection{Baselines}
Since, to the best of our knowledge, there doesn't exist a method designed to incorporate iterative feedback gathered over multiple rounds, we compare the proposed method to standard Stable Diffusion models in the following manner.
First, we run Stable Diffusion 1.5 (using the base model\footnote{\href{https://huggingface.co/runwayml/stable-diffusion-v1-5}{https://huggingface.co/runwayml/stable-diffusion-v1-5}} or a fine-tuned version called \emph{Dreamlike Photoreal}\footnote{\href{https://huggingface.co/dreamlike-art/dreamlike-photoreal-2.0}{https://huggingface.co/dreamlike-art/dreamlike-photoreal-2.0}} with or without HPS LoRA checkpoints) $N$ times, each with a different seed, generating $N$ batches of images. Then, we collect the desired evaluation metrics for the generated batch at each round and we use this values to perform quantitative comparisons with \name{}. 

It is important to note that while we may add images from each round as feedback to future rounds for our method, the baselines do not have a mechanism to incorporate feedback into future generations. Therefore, the baseline models generate images independently in each round without taking previous rounds into consideration. An overview of the all models taken into account can be found in Table \ref{tab:model_overview}.

\begin{table}[h]
    \centering
    \begin{tabular}{l c c c c}
        \toprule
         Name & SD version & Checkpoint & LoRA & FABRIC \\
         \midrule
         Baseline & $1.5$ & stable-diffusion-v1.5 & \xmark{} & \xmark{} \\
         Dreamlike Photoreal & $1.5$ & dreamlike-photoreal-2.0 & \xmark{} & \xmark{} \\
         HPS LoRA & $1.5$ & dreamlike-photoreal-2.0 & HPS LoRA & \xmark{} \\
         \midrule
         \name{} (SD) & $1.5$ & stable-diffusion-v1.5 & \xmark{} & \cmark{} \\
         \name{} & $1.5$ & dreamlike-photoreal-2.0 & \xmark{} & \cmark{} \\
         \name{} + HPS LoRA & $1.5$ & dreamlike-photoreal-2.0 & HPS LoRA & \cmark{} \\
         
         \bottomrule
    \end{tabular}
    \vspace{2mm}
    \caption{Specifics of \name{} models and baselines used during evaluation}
    \label{tab:model_overview}
\end{table}

\subsection{Metrics}

\subsubsection{Preference Model}
In order to automatically evaluate the experiments we use the PickScore  introduced in Section \ref{sec - related work} as a proxy score for general human preference. 

\subsubsection{CLIP Similarity to Feedback Images}
To assess the effectiveness of incorporating feedback into the generation process, we compute the CLIP similarity between the generated and the previous positive and negative feedback images.

For a generated image $x$ we compute the average CLIP-similarity to feedback images $y_\mathrm{pos}^{(1)}, \dots, y_\mathrm{pos}^{(k)}$ or $y_\mathrm{neg}^{(1)}, \dots, y_\mathrm{neg}^{(k)}$.
Specifically, let $\mathrm{CLIP}(x, y)$ denote the cosine similarity between CLIP-embeddings of $x$ and $y$. Then the positive CLIP similarity is defined as follows:
\[
s_\mathrm{pos}(x) = \frac{1}{k} \sum_{i=1}^k \mathrm{CLIP}(x, y_\mathrm{pos}^{(i)})
\]
The negative similarity $s_\mathrm{neg}(x)$ is defined analogously.

\subsubsection{In-batch Image Diversity}
In the process of steering the generation of images toward user preferences 
it is crucial to balance exploration (offering diverse image options for user selection) against exploitation (generating images aligned with previous feedback). To quantify the trade-off between these two factors across different rounds, we introduce a metric called In-batch Image Diversity.

For a batch of images $x_1, \dots, x_n$, we define the in-batch diversity based on the average CLIP-similarity between images in the batch. Specifically, with CLIP similarity defined as above, the In-batch Image Diversity $d$ is defined as follows:
\[
d(x_1, \dots, x_n) = 1 - \frac{2}{n(n-1)} \sum_{i=2}^n\sum_{j=1}^{i-1} \mathrm{CLIP}(x_i, x_j)
\]

where $\frac{n(n-1)}{2}$ is the number of elements in the upper triangular cosine similarity matrix. 


\section{Experiments}
In order to evaluate the capabilities of our model, we designed two experimental settings, each employing a different criterion for selecting feedback images during rounds.

In Sec.~\ref{sub:preference_model} we present a Preference Model-Based approach, where we select liked and disliked images using a preference score. In Sec.~\ref{sub:target_image} we present a Target Image-Based approach, where feedback selection relies on the similarity with a target image provided at the start of the experiment (but not used as feedback directly).

Both the experiments use a batch size of $4$ and consist on $3$ rounds of generation. After each round one image is selected as liked and another one as disliked. To demonstrate robustness with respect to the feedback strength, the preference model-based setup uses a feedback strength of $w = 0.1$ while the target image-based setup uses $w = 0.8$.

\subsection{Preference Model-based Feedback Selection}
\label{sub:preference_model}

The first experiment aimed to evaluate FABRIC by assessing how humans generally perceive generated images through an universal preference score. Therefore, this experiment operates under the assumption that all humans have the same preference when providing feedback, which in general may not always hold and is challenged in the second experiment. 

In practice, the experiment is conducted as follows.
First, a random set of $1000$ prompts is sampled from the HPS dataset \citep{hps}.
Next, for each prompt, after having initialized the user's liked and disliked images as empty sets, we simulate a 3-round interaction between the model and a user. In each round, the following steps are performed. First, FABRIC is run with the prompt and feedback images as input, generating a batch of $4$ images. Then, the Human Preference Score of each generated image is computed. Using those scores as a proxy for human feedback, the generated image with the highest score is added to the set of liked images, and the one with the lowest score is added to the disliked. 
For each batch of generated images we measure the average PickScore as well as the average CLIP similarity to both positive and negatives images. We compare each FABRIC model (dreamlike-photoreal-2.0 and HPS LoRA) against their respective baselines in terms of these two scores. 

\begin{figure}[t]    
    \centering
    \begin{subfigure}[t]{0.32\textwidth}
        \centering
        \includegraphics[width=\textwidth]{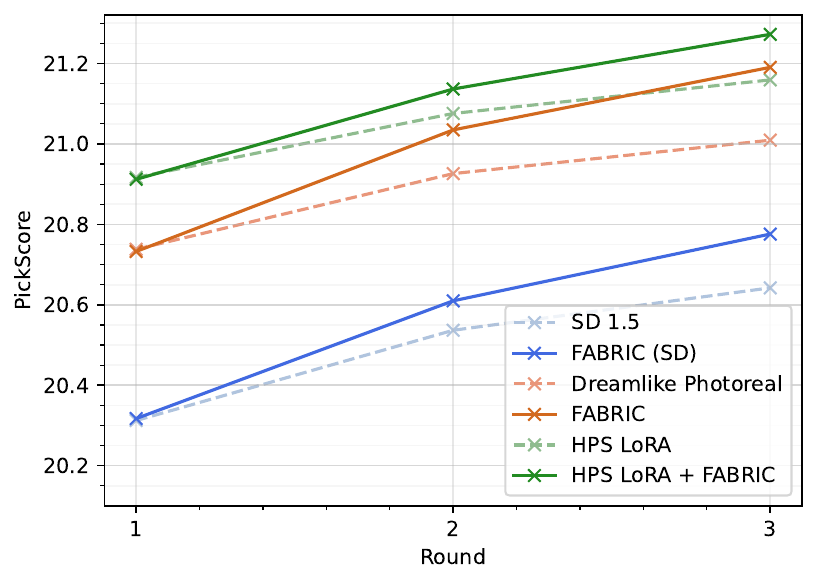}
        \caption{Highest PickScore of a generated image over all previous rounds.}
        \label{fig:preference-sim-a}
    \end{subfigure}
    \hfill
    \begin{subfigure}[t]{0.32\textwidth}
        \centering
        \includegraphics[width=\textwidth]{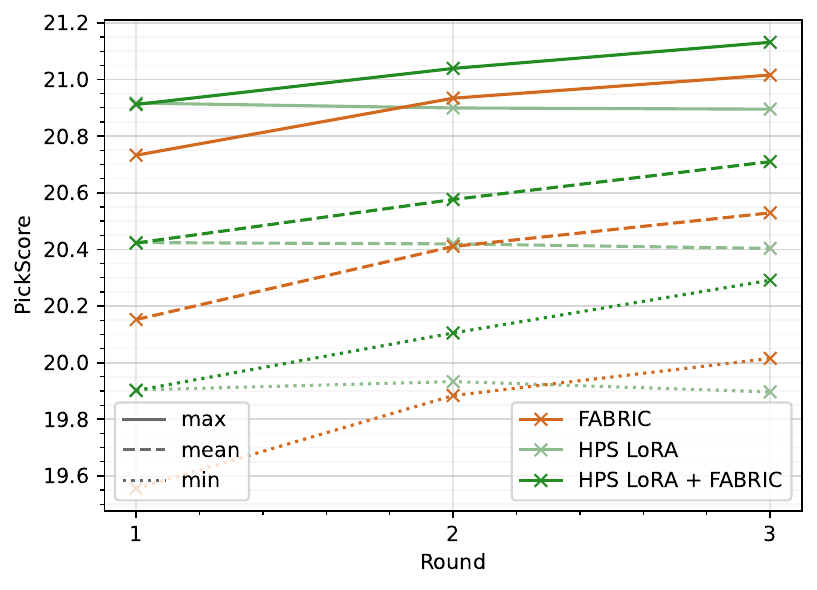}
        \caption{Min. (dotted), mean (dashed) and max. (solid) PickScore in each round.}
        \label{fig:preference-sim-b}
    \end{subfigure}
    \hfill
    \begin{subfigure}[t]{0.32\textwidth}
        \centering
        \includegraphics[width=\textwidth]{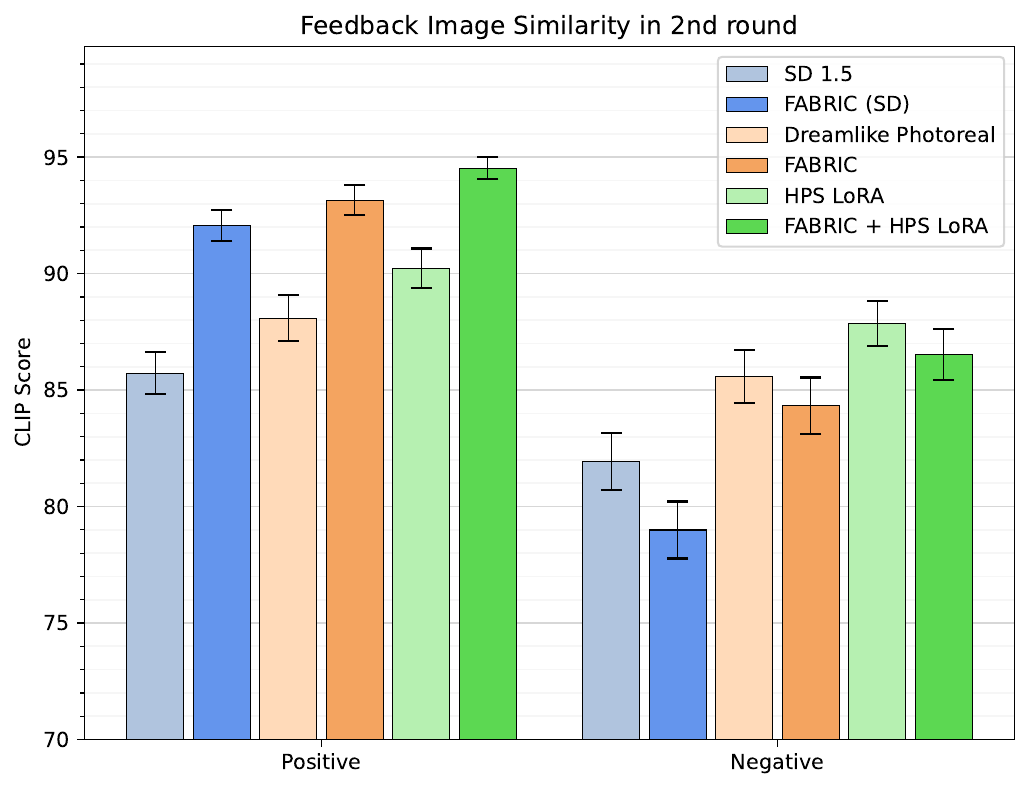}
        \caption{Similarity of generated images to positive/negative feedback.}
        \label{fig:preference-sim-c}
    \end{subfigure}
    \caption{Results of preference model-based feedback selection.}
    \label{fig:preference-sim}
\end{figure}

The results from our experiments are depicted in Figure~\ref{fig:preference-sim}. In Figure~\ref{fig:preference-sim-a} we address the concern that simply running multiple rounds and achieving higher image variety may lead to an increase in the maximum PickScore. Indeed we show that, even if in general this value increases over rounds, our model outperforms both the baselines. 
In Figure~\ref{fig:preference-sim-b} we observe that starting from the second round our model outperforms the baselines not only in terms of maximum PickScore but also in terms average and minimum value, indicating an overall enhancement in the quality of generated images. In Figure~\ref{fig:preference-sim-c} we evaluate the similarity of generated images to positive and negative feedback. We observe that even after just one round, the CLIP similarity score is higher for the positive samples, and lower for the negatives, compared to the baseline. This confirms that FABRIC effectively conditions the direction of generation based on the provided feedback.

\subsection{Target Image-based Feedback Selection}
\label{sub:target_image}

\begin{figure}[t]
    \centering
    \includegraphics[width=1.0\textwidth]{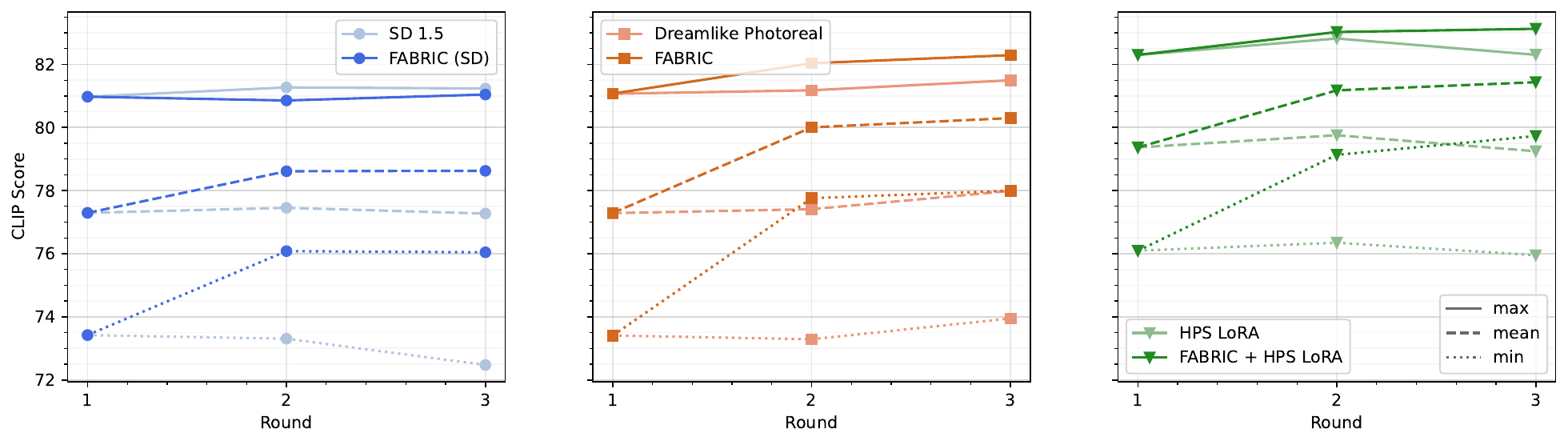}
    \vspace{-0.5cm}
    \caption{Results of target image-based feedback selection. Positive feedback improves target similarity over the baseline and using both positive and negative feedback improves it even further. At the same time, any kind of feedback reduces the diversity of generated images drastically.}
    \label{fig:target-sim}
\end{figure}

In this experiment, we challenge the assumption that all humans have the same preference, according to which they select liked/disliked images.
Instead, we assume that the user has some target image in mind, some imagined picture that they would like to express. 

To reflect this scenario, we manually gather a dataset of prompt-image pairs from the AI-art-sharing platform \href{https://prompthero.com/}{\textit{prompthero.com}} where users post their favorite generations along with the prompt that was used to generate the image. Due to this survivorship bias, we claim that it's appropriate to assume that shared images express the respective user's creative vision for the given prompt or, more generally, correspond to the user's preference.
The dataset is provided in our public GitHub repository.\footnote{\href{\githublink}{\githublink}}

During the experiment, we generate a batch of images using a prompt from the dataset and select feedback images based on CLIP-similarity to the associated target image: the most and least similar image is selected as positive and negative feedback respectively.
Note that the images in the dataset are generated with various different models and settings, which prevents the evaluated model from generating exact replications of the target image.

We compare the SD-1.5 and Dreamlike Photoreal baselines (using no feedback) to regular FABRIC in terms of similarity to the specified target image and in terms of in-batch image diversity.
The results of this are illustrated in Figure~\ref{fig:target-sim}.
We find that both of them outperform the respective baselines, improving both best-case and worst-case outcomes in rounds 2 and 3.
Especially the per-round minimum similarity drastically improves over the baseline when any kind of feedback is introduces.

Qualitative results from this experiment are shown in Appendix~\ref{sec - target image based examples}.

\subsection{Adapting the Feedback Schedule}
\label{sec - adapting the feedback schedule}
In addition to the experiments reported in the main part, we investigated the adaptation of the feedback schedule. The default configuration of FABRIC adds feedback in every denoising step. Our findings show that restricting feedback in the first half of the denoising process improves performance by a large margin. On the other hand, only including feedback in the second half of the denoising process reduces performance. This leads to the hypothesis that feedback is useful in the early stages of the denoising process, but fine-grained details are largely determined by the prompt, where feedback does not help. 

\subsection{Increasing Diversity with Prompt Dropout}

\begin{figure}[t]
    \centering
    \begin{subfigure}[t]{0.4\textwidth}
        \centering
        \includegraphics[width=\textwidth]{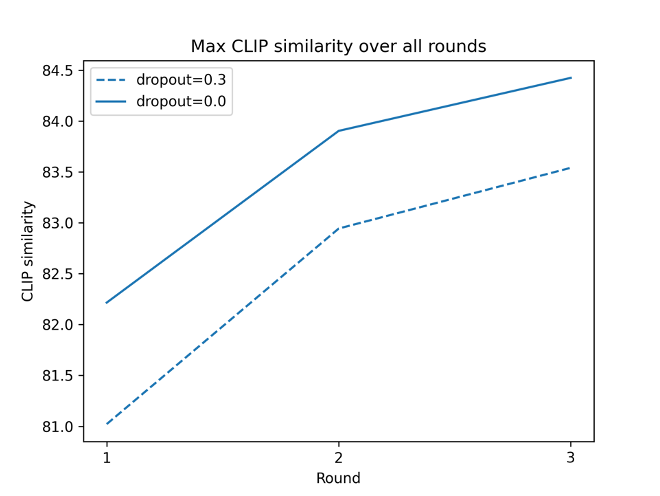}
        \caption{The effect of prompt dropout on the best CLIP similarity to the target image.}
        \label{fig:prompt-dropout-a}
    \end{subfigure}
    \hspace{0.5cm}
    \begin{subfigure}[t]{0.4\textwidth}
        \centering
        \includegraphics[width=\textwidth]{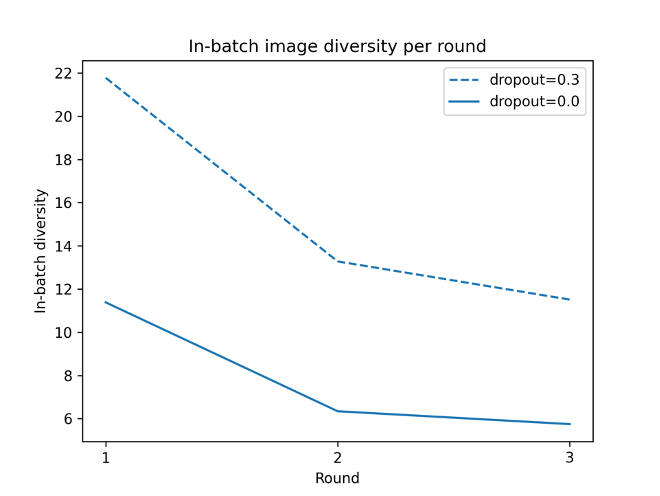}
        \caption{The effect of prompt dropout on the in-batch image diversity.}
        \label{fig:prompt-dropout-b}
    \end{subfigure}
    \caption{Prompt dropout appears to be an effective way of trading CLIP similarity for more diversity in the generative distribution.}
    \label{fig:prompt-dropout}
\end{figure}

As can be seen in Appendix~\ref{sec - target image based examples}, the diversity of generated images quickly collapses as soon as any kind of feedback is introduced.
This is likely due to the conditioning mechanism, which pushes the modal towards generating images very similar to the reference.
From a user perspective, however, this drop in diversity is undesirable, as it prevents future iterations from discovering new and possibly better results.

We investigate one possible approach to combatting this collapse by using prompt-dropout, i.e. dropping every word in the prompt with some probability.
Using the same setup as in Section~\ref{sub:target_image}, we compare the best CLIP similarity to the target as well as the in-batch image diversity of using prompt dropout with $p = 0.3$ against not using it ($p = 0.0$).
We find that there is a direct trade-off between CLIP similarity and diversity: $p=0.3$ significantly increases diversity but also produces worse images over all.
Note that an increased initial diversity especially will also increase diversity in feedback images, which may explain why even after 3 rounds, the diversity of $p=0.3$ is still above the initial diversity of $p=0.0$.
Thanks to more varied feedback, prompt dropout might be able to catch up to the baseline given enough feedback.

\section{Discussion}




\paragraph{Limitations}
While FABRIC works well in our experiments, some limitations apply.
We noticed that \name{} is effective at constraining the generative distribution to a subset of preferable outcomes, but it struggles to expand the distribution beyond the initial text-conditioned distribution given by the model.
Especially since feedback is already sampled from the model's output, additional modifications will be needed to overcome this shortcoming.
For the same reason, the diversity of generated images quickly collapses to a single mode close to the feedback images. We propose prompt dropout as a possible remedy, but this might run the risk of dropping crucial words in the prompt and changing the generations completely.

Another limitation lays in the way we collect feedback. Currently, we only allow users to provide binary preferences over images, which does not allow for specific conditioning in the image generation process.

\paragraph{Future Work}
Future work will investigate approaches toward increasing diversity or generally controlling the exploration-exploitation trade-off in a more principled fashion.
An interesting avenue for this could be to retrieve candidate images from an external image corpus (e.g., the user's previously liked images or a corpus of general high-quality images) and use them as a seed for the feedback loop. A big advantage of FABRIC is its orthogonality to most other stable diffusion variations, like checkpoints, LoRA weights, modifying or replacing the text conditioning (\cite{xu2023promptfree}) while achieving significant improvements on top of them (see FABRIC + LoRA). 
Ultimately, one could specify precise weights for each liked or disliked image and rate if rather coarse features or fine-grained features are liked or disliked, respectively. One potential improvement in this direction would be to specify whether user preferences are expressed based on the structure or style of the image. This distinction would allow for more specific conditioning in the image-generation process. 
In addition, FABRIC provides a well-defined action space with different parameters that can affect the generated results.
This opens up the avenue for performing Bayesian optimization on an arbitrary objective (e.g., direct user feedback or a score given by a preference model).

\section{Conclusion}

We present FABRIC, a training-free method that incorporates iterative feedback into the generation process of text-to-image models, leveraging attention-based reference image conditioning.
Our experimental findings suggest that FABRIC is capable of implicitly optimizing a variety of objective functions such as human preference and similarity to a designated target image.

These objectives noticeably improve with more feedback rounds, demonstrating that FABRIC's effectiveness significantly exceeds merely sampling more images without providing additional feedback. Remarkably, even without training or hyperparameter tuning, FABRIC can outperform the HPS LoRA, a model explicitly trained to optimize human preference, on the relevant metric.

Notably, FABRIC tends to trade exploration for exploitation, often collapsing to a uniform distribution after a handful of feedback rounds. We have examined a few strategies to alleviate this collapse, though further investigation of this trade-off remains a prospect for future work.

The iterative setting is paramount to how generative visual models are used in practice. Despite this, recent research on aligning text-to-image models has left it largely unexplored.
We believe that this study contributes towards the formation of a framework that aids in devising and evaluating methodologies intended to address this setting.

\section*{Ethical Implications}

Text-to-image models have the potential to make creative visual expression more accessible to everyone by allowing individuals without artistic skills or technical expertise to generate visually appealing content. Our proposed method aims to further enhance the accessibility and personalization of these models by incorporating user preferences into the image generation process. This is achieved through the utilization of positive and negative feedback, allowing users to provide natural and intuitive guidance based on prior images or previously generated ones.

By adopting our approach, users gain increased control over the generated content, promoting ethical usage. However, this heightened control also raises concerns regarding the potential for misuse or abuse of the system. To address these concerns, it becomes crucial for the community as well as society at large to establish clear guidelines regarding the legal and ethical utilization of such systems.
By placing responsibility on individual users to ensure responsible and ethical usage, we can mitigate the risks and foster a positive and constructive environment for creative expression.

\medskip

\medskip
\bibliographystyle{iclr2022_conference}
\bibliography{iclr2022_conference}

\appendix

\section{Appendix}\label{sec:appendix}

\subsection{Method Details}
\label{sec - pseudocode}

Here we provide the Pseudocode of the FABRIC algorithm (see Algorithm \ref{algorithm - fabric}). 

\begin{equation}
    \label{eq:weightedattention}
    \mathrm{WeightedAttention}_w(Q, K, V) = \left(
    \frac{w}{\norm{w}_1} 
    \odot 
    \mathrm{softmax}\left(
    \frac{QK^\top}{\sqrt{d_k}}
    \right)\right)V^\top
\end{equation}

\begin{algorithm}
    \caption{FABRIC: Feedback via Attention-Based Reference Image Conditioning}
    \begin{algorithmic}[1]
        \label{algorithm - fabric}

        \Require Let $N$ be the number of feedback rounds, $n$ be the batch size of generated images in each round and \textit{model} be a diffusion model capable of reference-conditioning.
        \Procedure{FABRIC}{}
            \State pos, neg $\gets$ [], []
            \For{$i \in \{1, \dots, N\}$}
                \State prompt $\gets$ get\_prompt($i$)
                \State images $\gets$ [\Call{Generate}{prompt, pos, neg} \textbf{for} $n$ \textbf{times}]
                \State $x_\mathrm{pos}$, $x_\mathrm{neg}$ $\gets$ get\_feedback(images) \Comment{we focus on one like and one dislike}
                \State pos.put($x_\mathrm{pos}$)
                \State neg.put($x_\mathrm{neg}$)
            \EndFor
        \EndProcedure
        \Function{Generate}{prompt, positives, negatives}
            \State $z_T$ = initial\_noise()
            \For{$t \in \{T, \dots, 1\}$}
                \State hiddens $\gets$ \{\}
                \For{$x_\mathrm{ref} \in \{\dots\text{positives},\; \dots\text{negatives}\}$}
                    \State $z_\mathrm{ref}$ $\gets$ $\sqrt{\bar{\alpha}_t} \cdot x_\mathrm{ref} + \sqrt{1 - \bar{\alpha}_t} \cdot \epsilon_\mathrm{ref}^{(t)}$ \Comment forward diffusion noising, $\epsilon_\mathrm{ref}^{(t)} \sim \mathcal{N}(0, \mathbf{I})$
                    \State $h$ $\gets$ \Call{PrecomputeHiddenStates}{$z_\mathrm{ref}$, $t$}
                    \State hiddens.put($h$)
                \EndFor
                \State compute $w_\mathrm{pos}^{(t)}$ and $w_\mathrm{neg}^{(t)}$ according to the feedback settings
                \State $\epsilon_{\mathrm{cond},t-1}$ $\gets$ \Call{ModifiedUnet}{$z_t$, $t$, get\_positive(hiddens), $w_\mathrm{pos}^{(t)}$}
                \State $\epsilon_{\mathrm{uncond},t-1}$ $\gets$ \Call{ModifiedUnet}{$z_t$, $t$, get\_negative(hiddens), $w_\mathrm{neg}^{(t)}$}
                \State $z_{t-1}$ $\gets$ step\_with\_cfg($z_t$, $\epsilon_{\mathrm{cond},t}$, $\epsilon_{\mathrm{uncond},t}$) \Comment{ancestral Euler sampling in our case}
            \EndFor
            \State \Return $z_0$
        \EndFunction
        \Function{PrecomputeHiddenStates}{$z$, $t$}
            \State hiddens $\gets$ []
            \For{$i$-th layer in the Unet}
                \State apply ResNet block(s) to $z$
                \State hiddens.put(i, $z$) \Comment{just before self-attention}
                \State apply self-attention to $z$
                \State apply cross-attention and FFN to $z$
            \EndFor
            \State \Return $z$
        \EndFunction
        \Function{ModifiedUnet}{$z$, $t$, hiddens, $w$}
            \For{$i$-th layer in the Unet}\Comment{for hiddens $=\emptyset$ this function is a standard U-Net}
                \State $h$ $\gets$ hiddens.at(i)
                \State apply ResNet block(s) to $z$
                \State $Q$ $\gets$ $W_Q^{(i)} \cdot z$
                \State $K$ $\gets$ $W_K^{(i)} \cdot \mathrm{concat}(z, h)$
                \State $V$ $\gets$ $W_V^{(i)} \cdot \mathrm{concat}(z, h)$
                \State $z$ $\gets$ $\mathrm{WeightedAttention}_{w}(Q, K, V)$
                \State apply cross-attention and FFN to $z$
            \EndFor
            \State \Return $z$
        \EndFunction
    \end{algorithmic}
\end{algorithm}

\subsection{Target Image-based Feedback Selection Examples}
\label{sec - target image based examples}

Here, we provide some example feedback-trajectories from the promthero-dataset, see Figure \ref{figure:target-experiment}


\begingroup
\setlength{\tabcolsep}{2pt}
\begin{figure}[]
    \centering
    \begin{tabular}{c|ccc}
        Target Image & Round 0 & Round 1 & Round 2 \\
        \includegraphics[width=2cm]{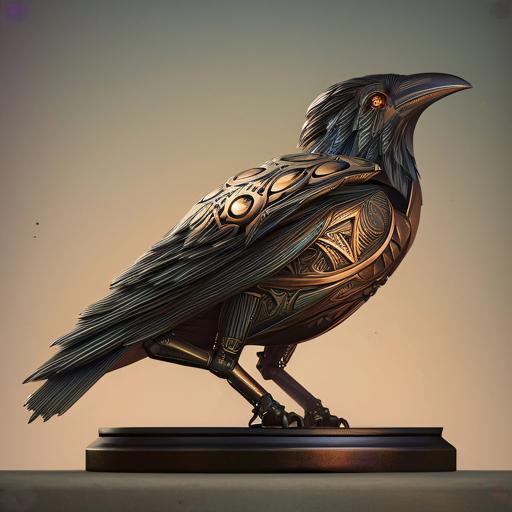} & \includegraphics[width=3.8cm]{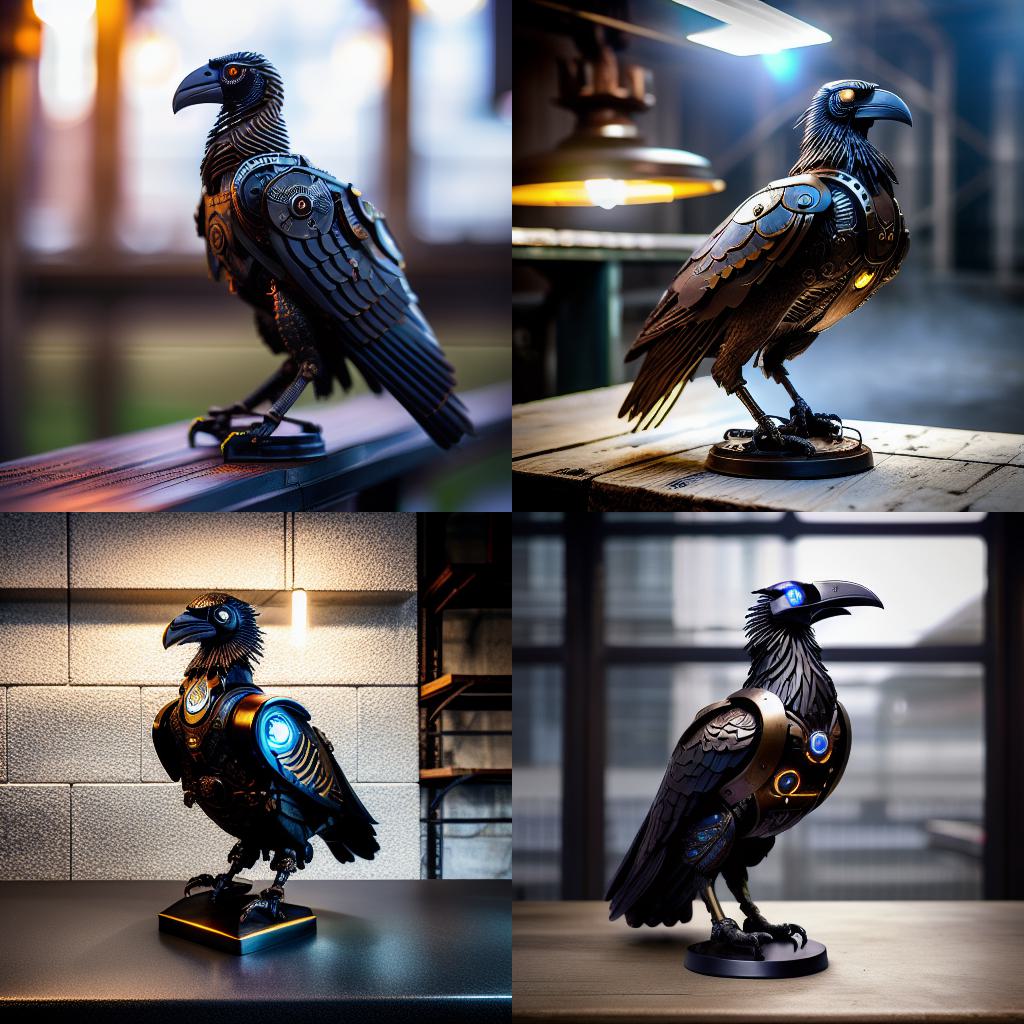} & \includegraphics[width=3.8cm]{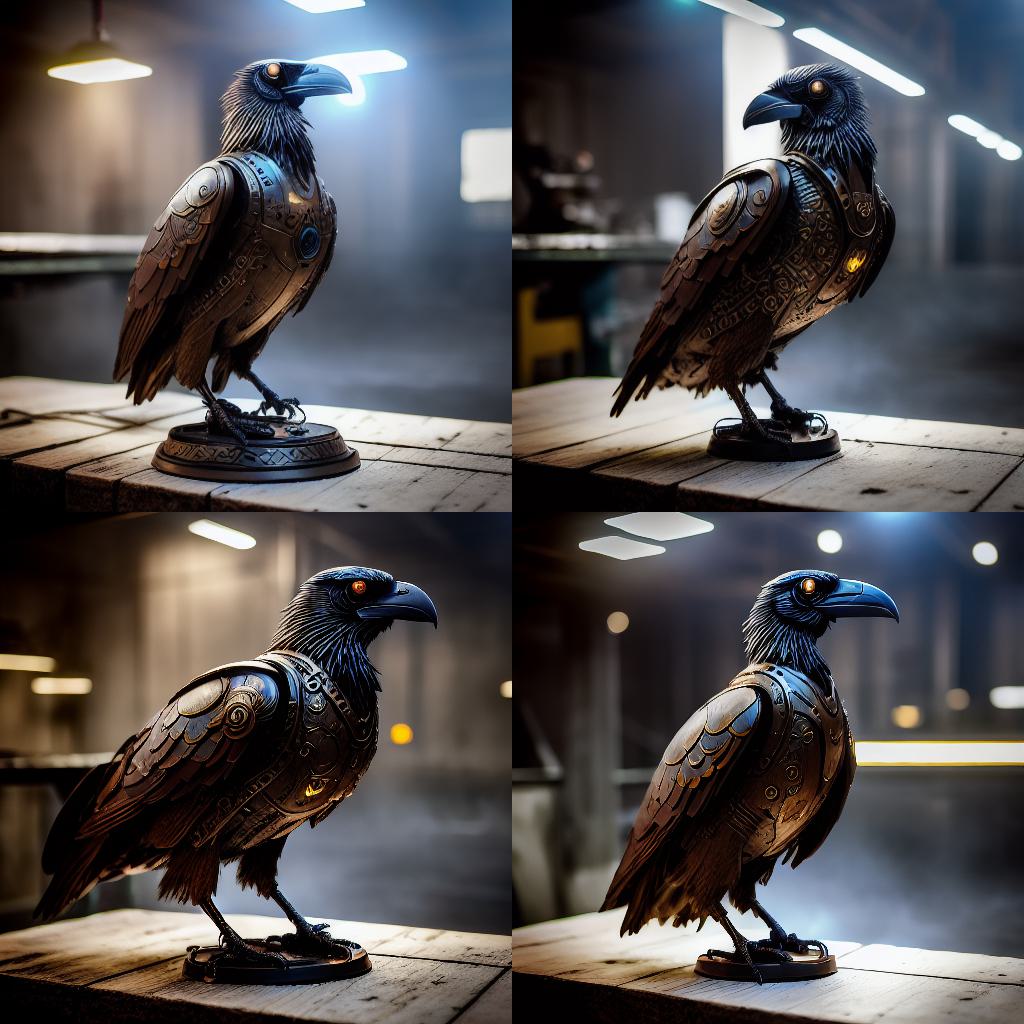} & \includegraphics[width=3.8cm]{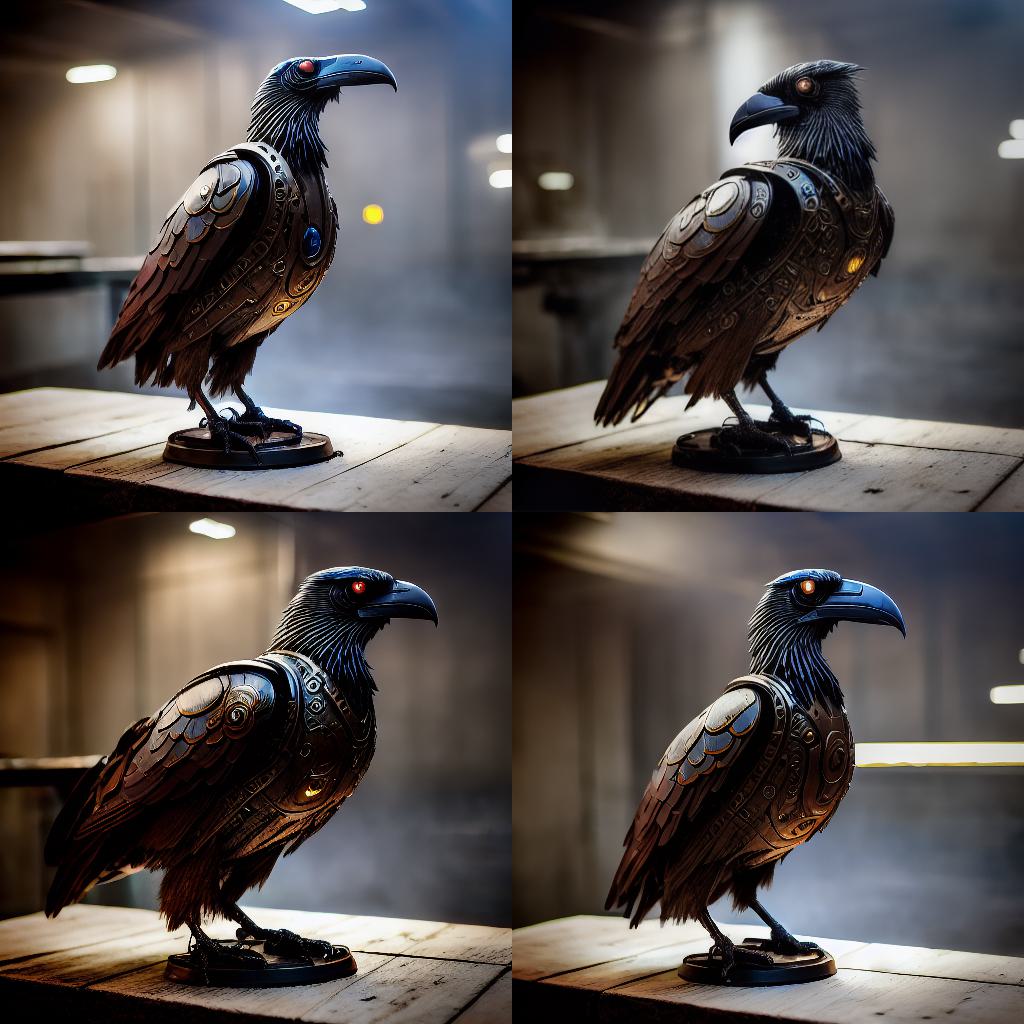} \\
        \includegraphics[width=2cm]{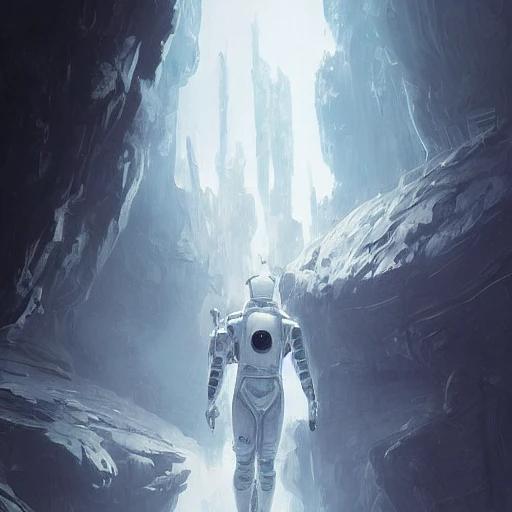} & \includegraphics[width=3.8cm]{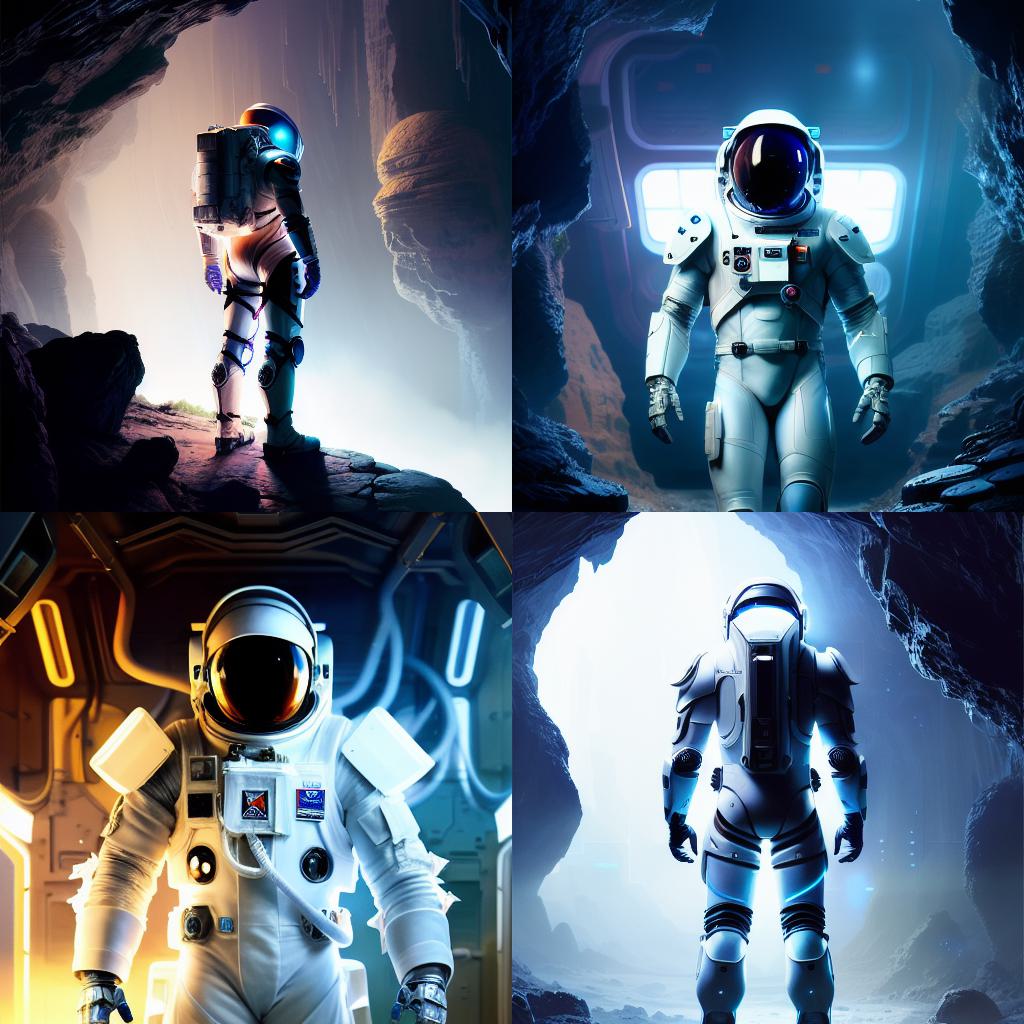} & \includegraphics[width=3.8cm]{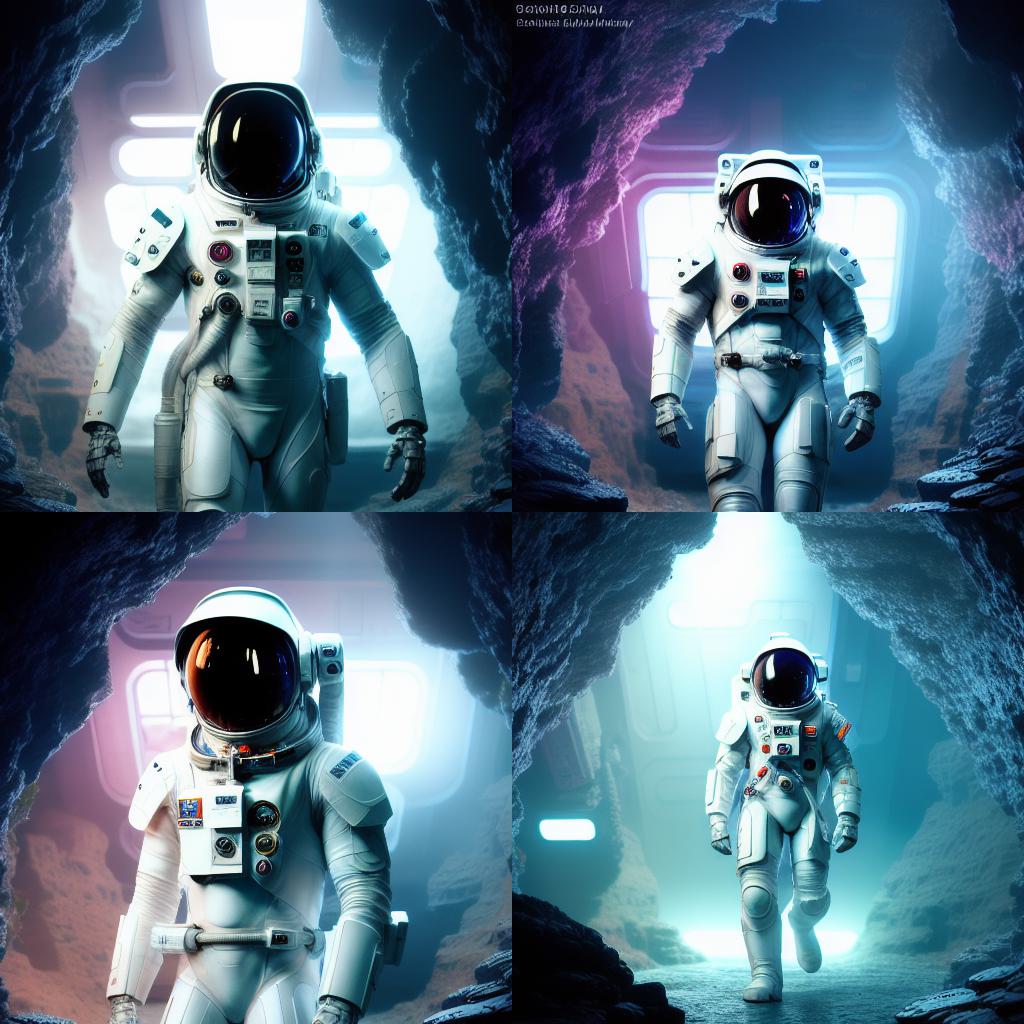} & \includegraphics[width=3.8cm]{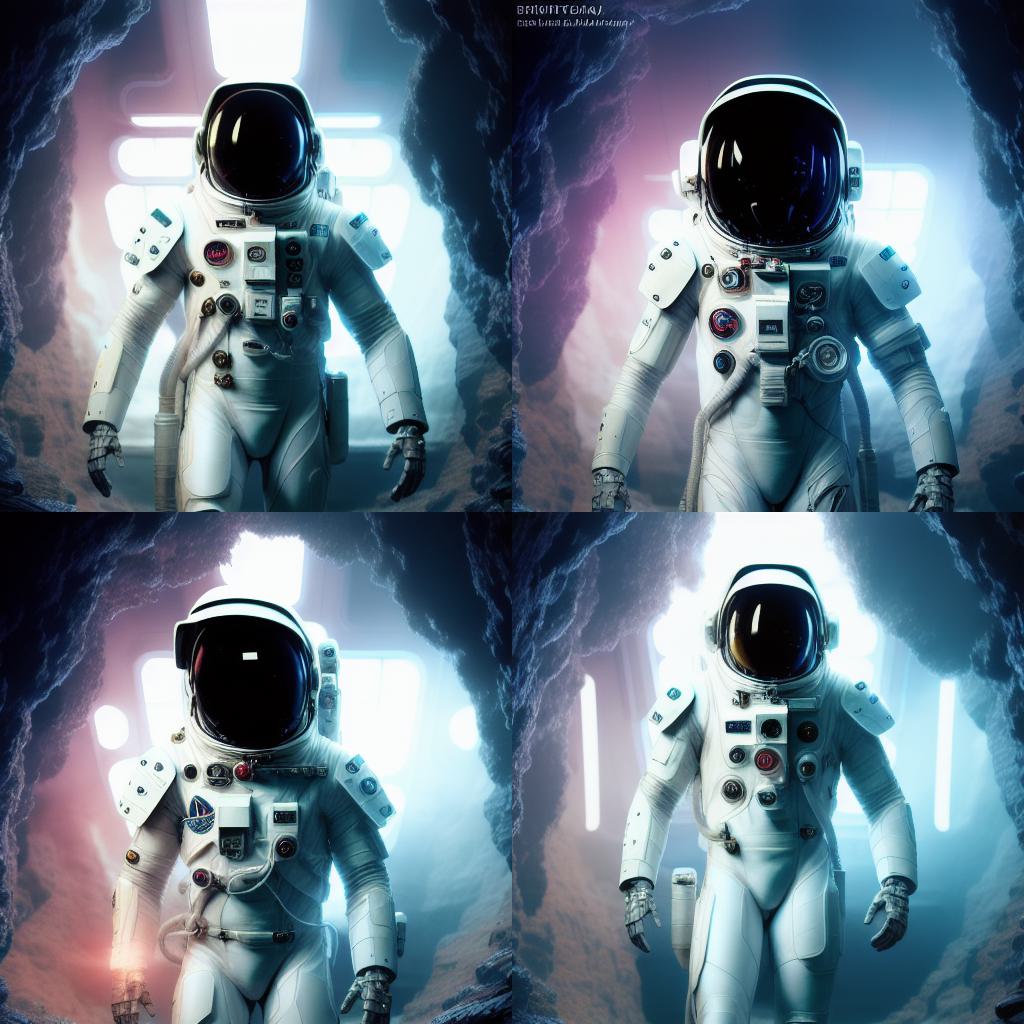} \\
        \includegraphics[width=2cm]{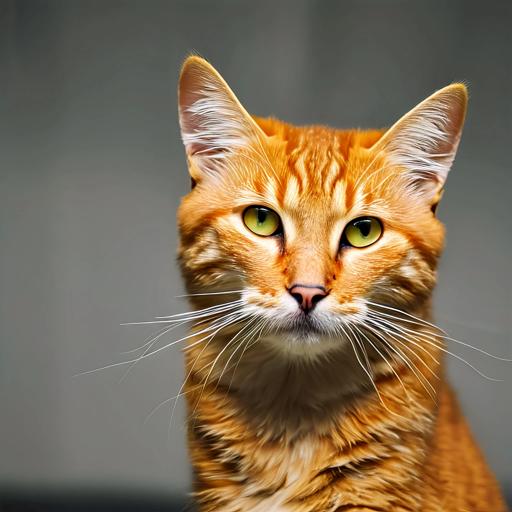} & \includegraphics[width=3.8cm]{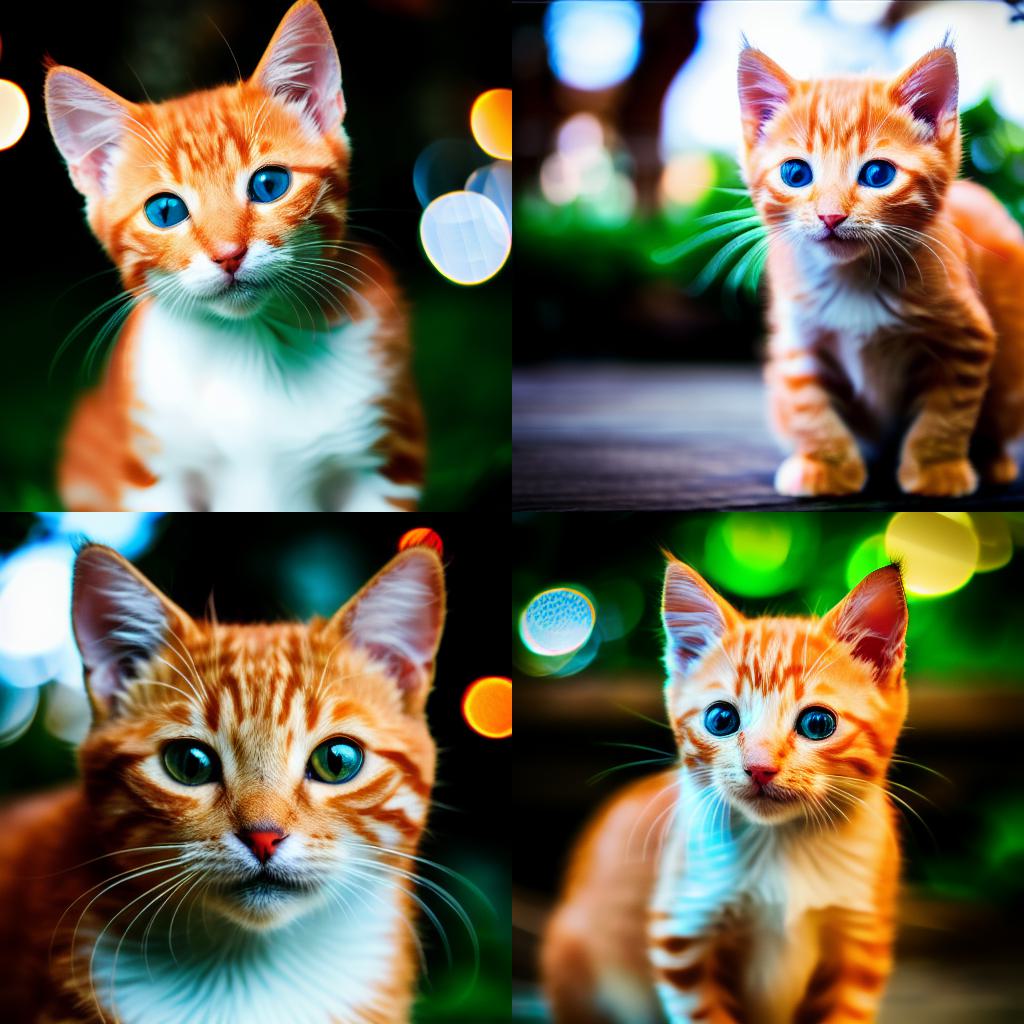} & \includegraphics[width=3.8cm]{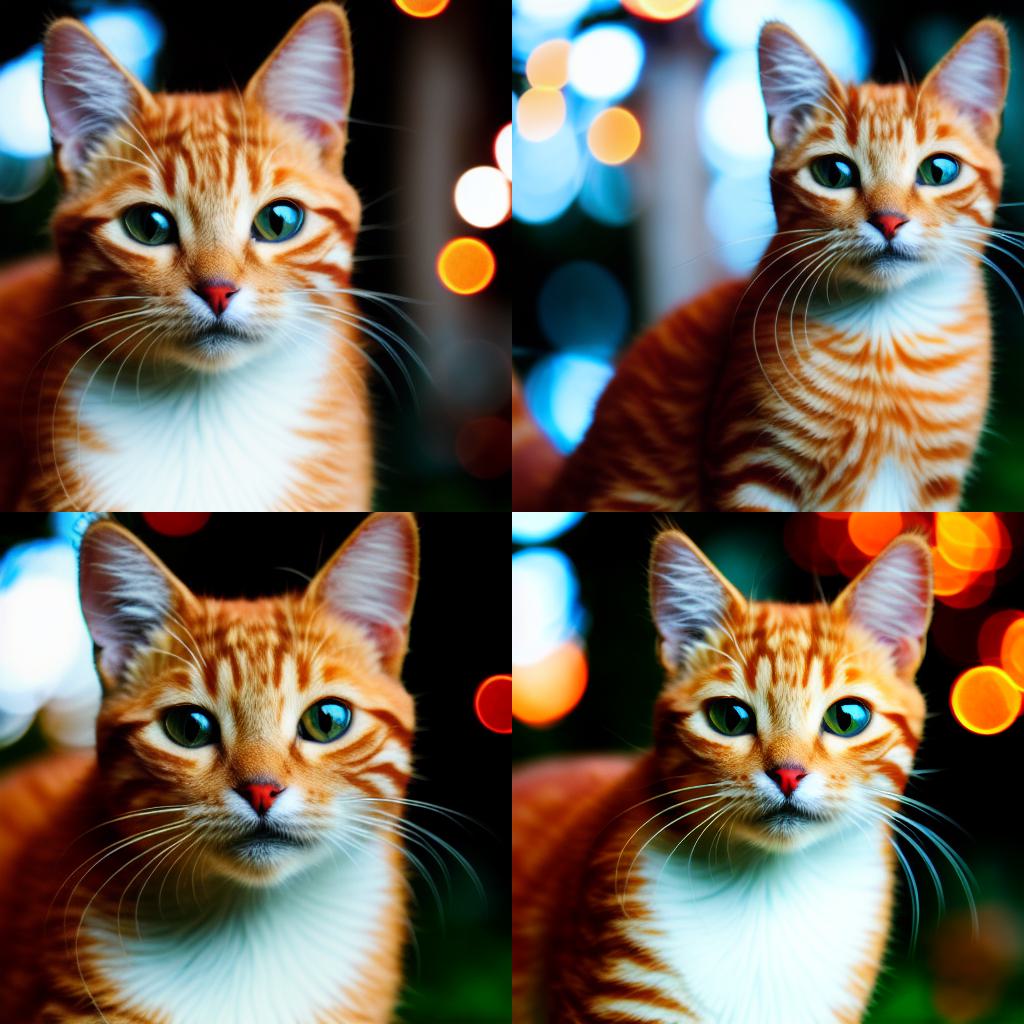} & \includegraphics[width=3.8cm]{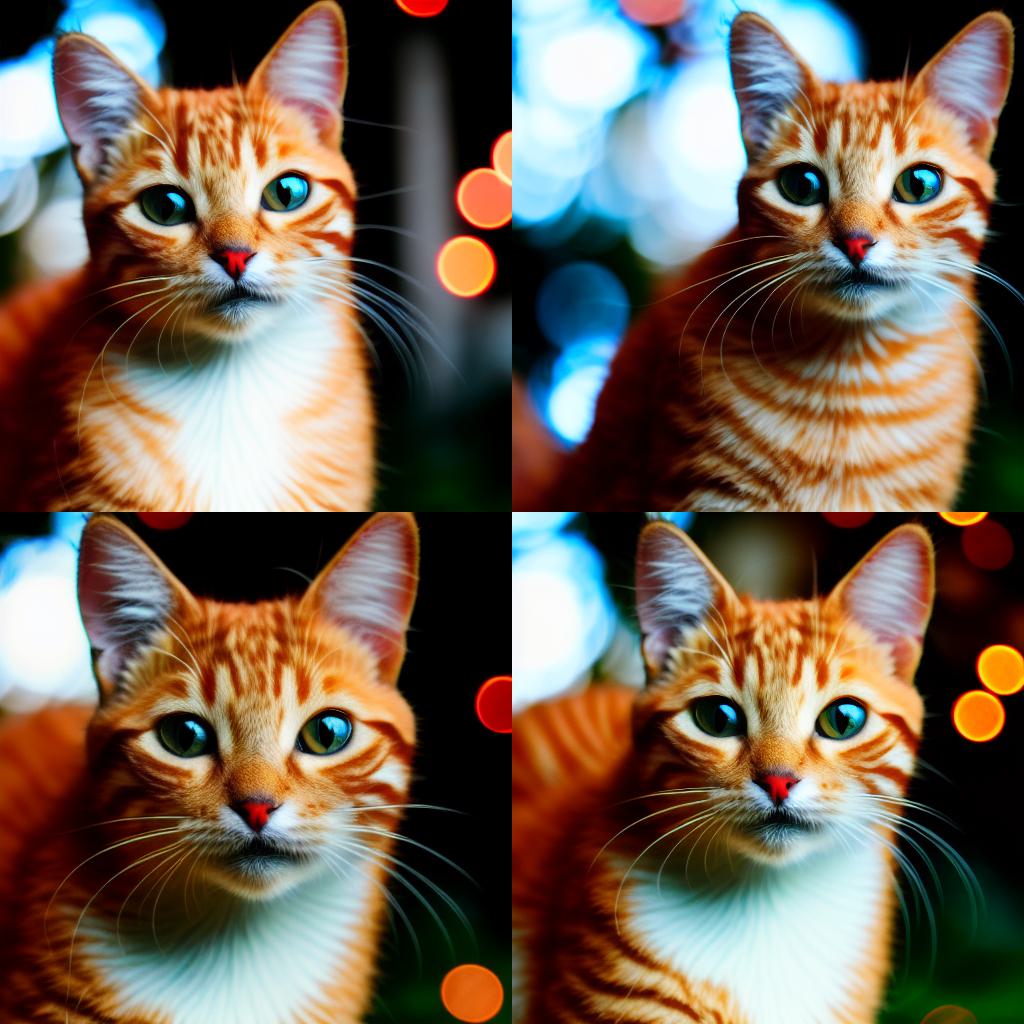} \\
        \includegraphics[width=2cm]{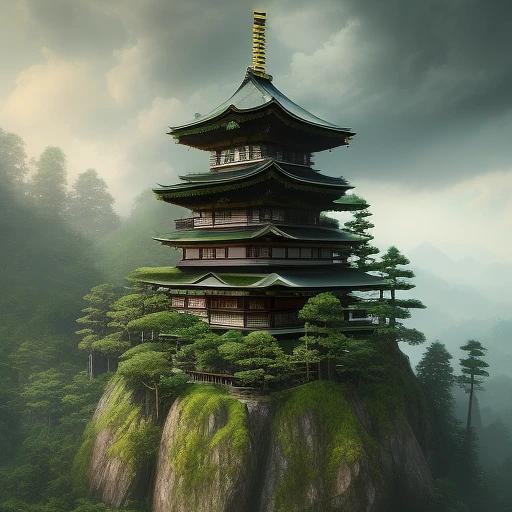} & \includegraphics[width=3.8cm]{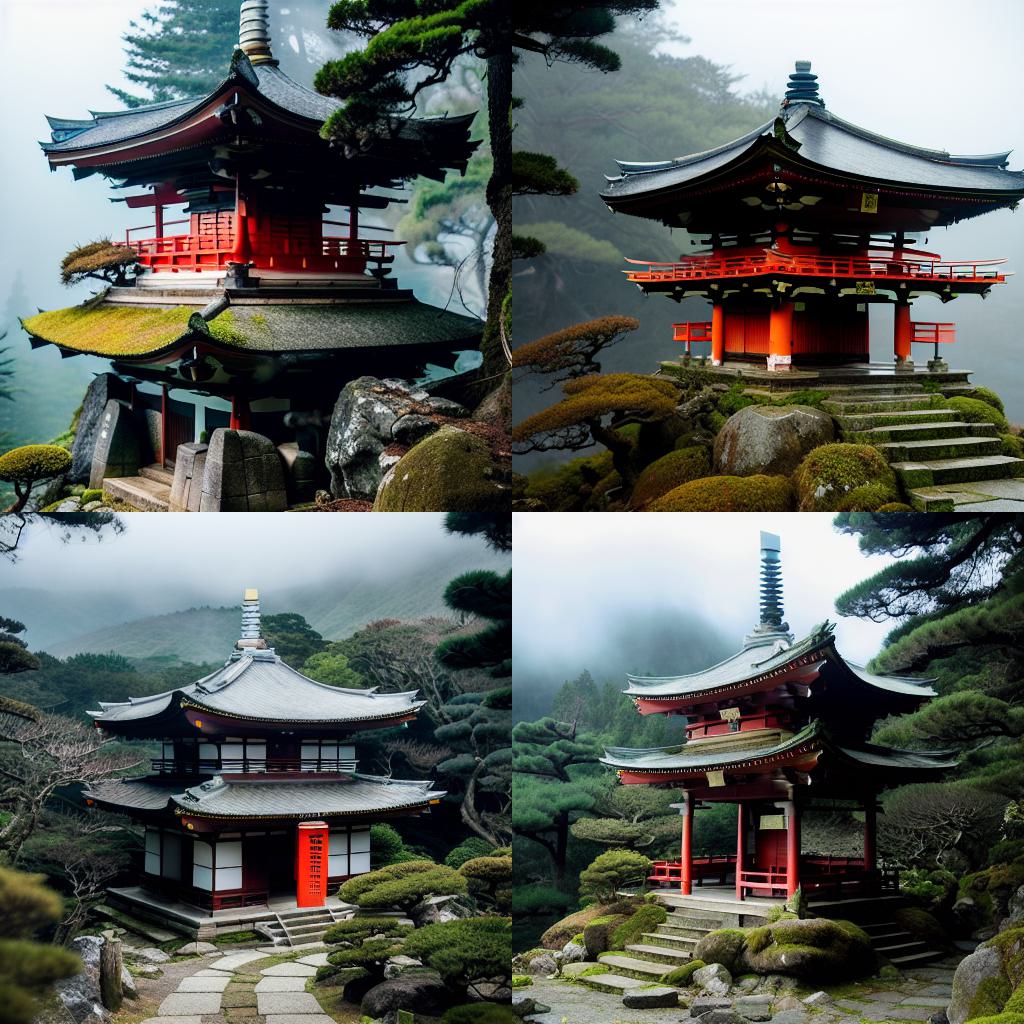} & \includegraphics[width=3.8cm]{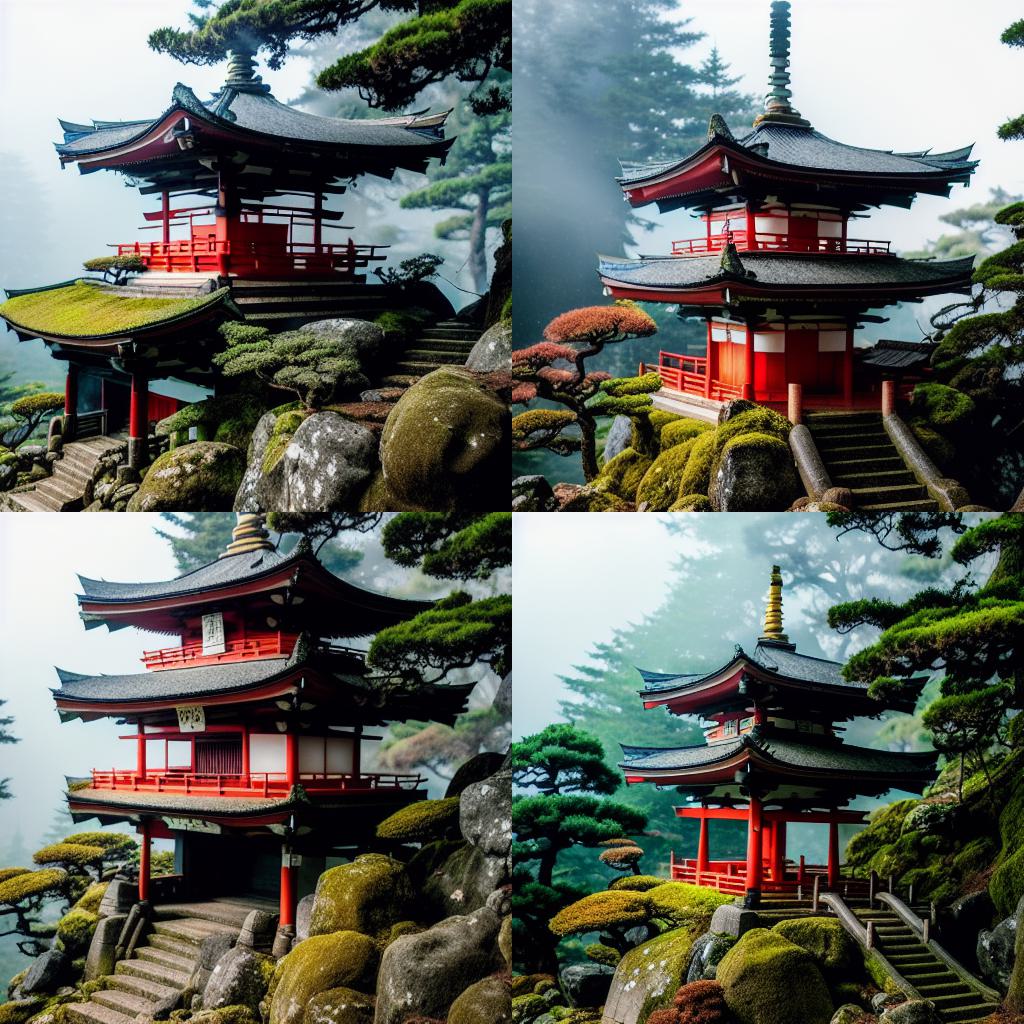} & \includegraphics[width=3.8cm]{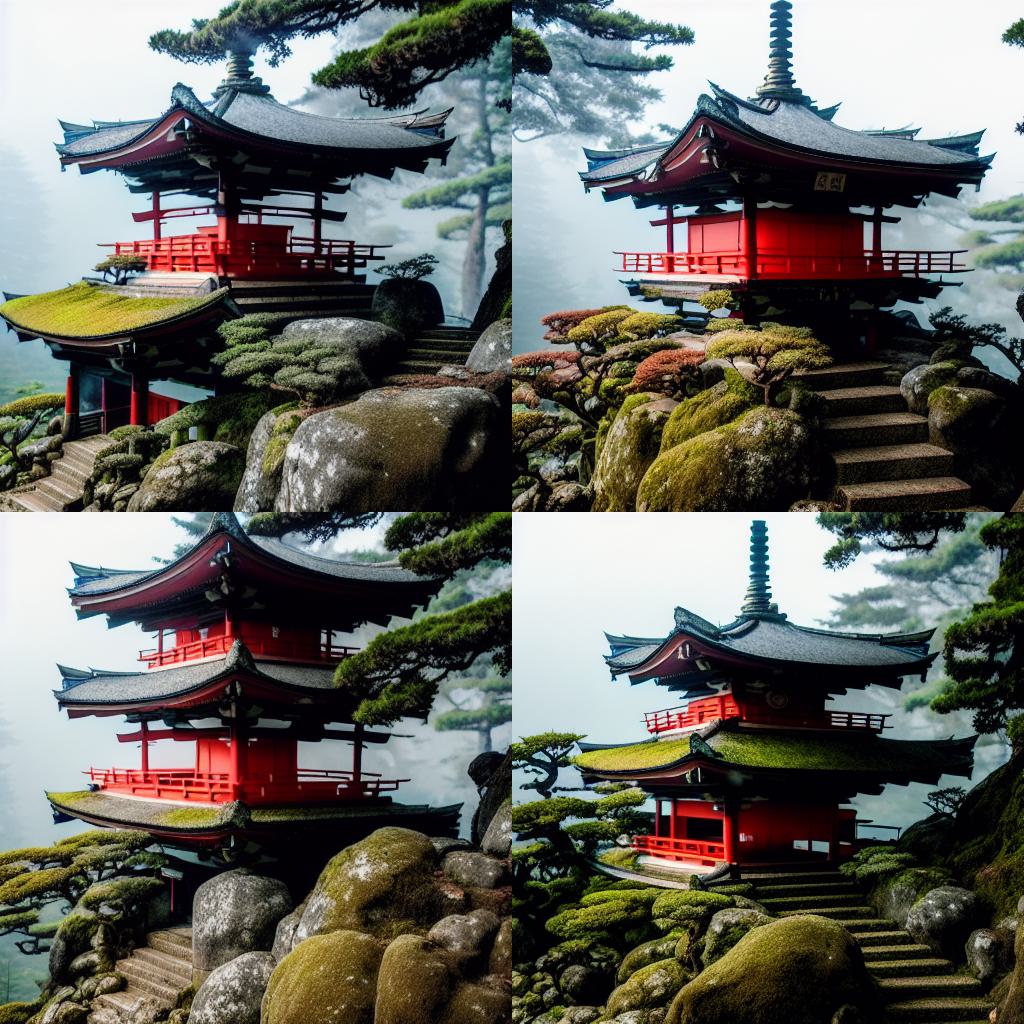} \\
    \end{tabular}
    \caption{Examples of feedback rounds from our target-image-based experiment.}
    \label{figure:target-experiment}
\end{figure}
\endgroup

\subsection{FABRIC Workflow}
\begin{figure}
    \centering
    \scalebox{0.25}      {\includegraphics{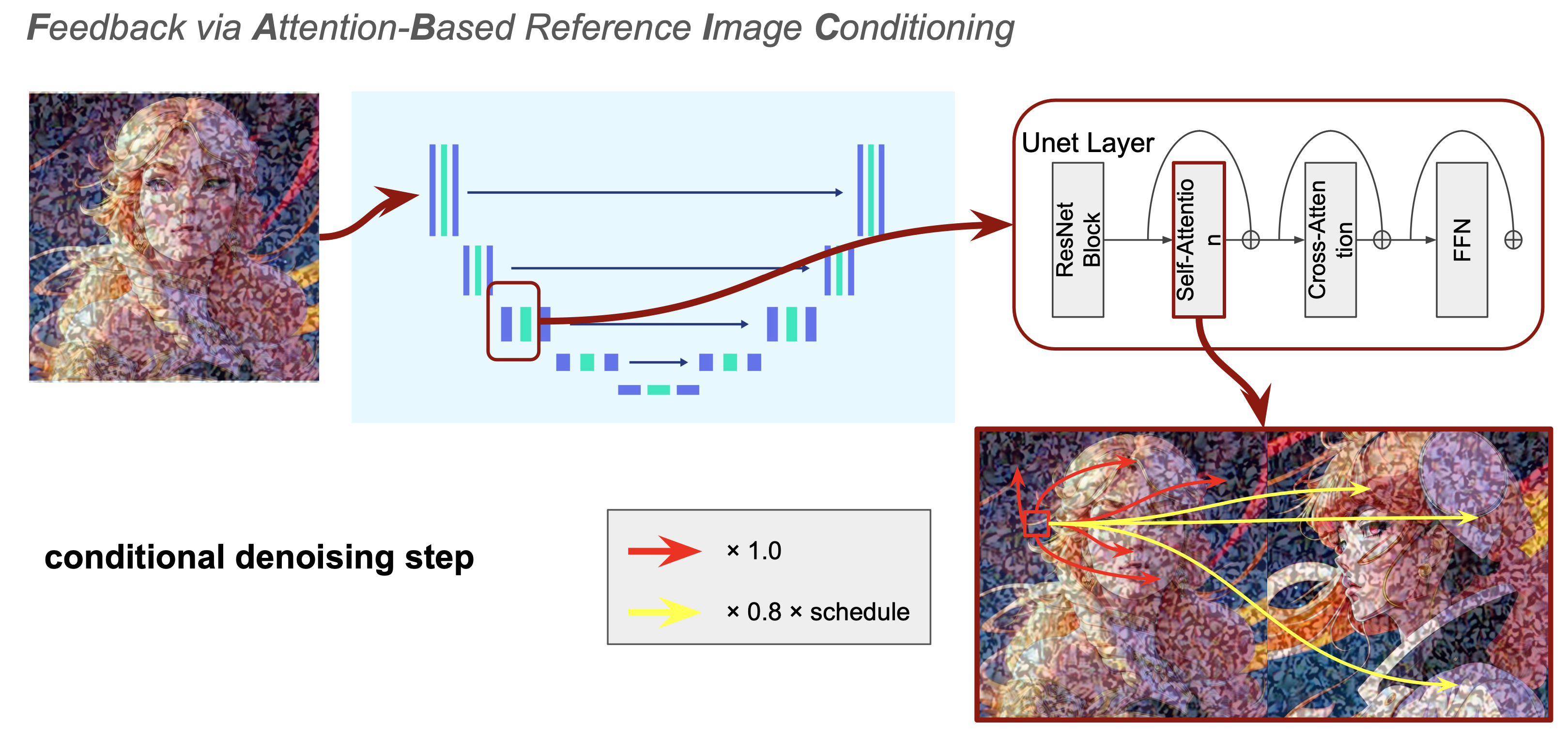}}
    \caption{FABRIC: Reference images are noised up to a certain step, and the extracted keys and values are injected into the self-attention of the U-Net during the denoising process.}
    \label{fig: FABRIC workflow}
\end{figure}

In Figure \ref{fig: FABRIC workflow} we show the high-level workflow of FABRIC.

\end{document}

%% file: math_commands.tex
\usepackage{amsmath,amsfonts,bm}

\def\eqref#1{equation~\ref{#1}}

\def\1{\bm{1}}

\DeclareMathAlphabet{\mathsfit}{\encodingdefault}{\sfdefault}{m}{sl}
\SetMathAlphabet{\mathsfit}{bold}{\encodingdefault}{\sfdefault}{bx}{n}













%% file: iclr2022_conference.bbl
\begin{thebibliography}{27}
\providecommand{\natexlab}[1]{#1}
\providecommand{\url}[1]{\texttt{#1}}
\expandafter\ifx\csname urlstyle\endcsname\relax
  \providecommand{\doi}[1]{doi: #1}\else
  \providecommand{\doi}{doi: \begingroup \urlstyle{rm}\Url}\fi

\bibitem[Brock et~al.(2019)Brock, Donahue, and Simonyan]{brock2019large}
Andrew Brock, Jeff Donahue, and Karen Simonyan.
\newblock Large scale gan training for high fidelity natural image synthesis,
  2019.

\bibitem[Chan et~al.(2020)Chan, Wang, Xu, Gu, and Loy]{chan2020glean}
Kelvin C.~K. Chan, Xintao Wang, Xiangyu Xu, Jinwei Gu, and Chen~Change Loy.
\newblock Glean: Generative latent bank for large-factor image
  super-resolution, 2020.

\bibitem[Dao et~al.(2022)Dao, Fu, Ermon, Rudra, and Ré]{dao2022flashattention}
Tri Dao, Daniel~Y. Fu, Stefano Ermon, Atri Rudra, and Christopher Ré.
\newblock Flashattention: Fast and memory-efficient exact attention with
  io-awareness, 2022.

\bibitem[Fan et~al.(2023)Fan, Watkins, Du, Liu, Ryu, Boutilier, Abbeel,
  Ghavamzadeh, Lee, and Lee]{fan2023dpok}
Ying Fan, Olivia Watkins, Yuqing Du, Hao Liu, Moonkyung Ryu, Craig Boutilier,
  Pieter Abbeel, Mohammad Ghavamzadeh, Kangwook Lee, and Kimin Lee.
\newblock Dpok: Reinforcement learning for fine-tuning text-to-image diffusion
  models, 2023.

\bibitem[Gal et~al.(2023)Gal, Alaluf, Atzmon, Patashnik, Bermano, Chechik, and
  Cohen-or]{gal2023an}
Rinon Gal, Yuval Alaluf, Yuval Atzmon, Or~Patashnik, Amit~Haim Bermano, Gal
  Chechik, and Daniel Cohen-or.
\newblock An image is worth one word: Personalizing text-to-image generation
  using textual inversion.
\newblock In \emph{The Eleventh International Conference on Learning
  Representations}, 2023.
\newblock URL \url{https://openreview.net/forum?id=NAQvF08TcyG}.

\bibitem[Goodfellow et~al.(2014)Goodfellow, Pouget-Abadie, Mirza, Xu,
  Warde-Farley, Ozair, Courville, and Bengio]{goodfellow2014generative}
Ian~J. Goodfellow, Jean Pouget-Abadie, Mehdi Mirza, Bing Xu, David
  Warde-Farley, Sherjil Ozair, Aaron Courville, and Yoshua Bengio.
\newblock Generative adversarial networks, 2014.

\bibitem[Ho \& Salimans(2022)Ho and Salimans]{ho2022classifierfree}
Jonathan Ho and Tim Salimans.
\newblock Classifier-free diffusion guidance, 2022.

\bibitem[Ho et~al.(2020)Ho, Jain, and Abbeel]{ho2020denoising}
Jonathan Ho, Ajay Jain, and Pieter Abbeel.
\newblock Denoising diffusion probabilistic models, 2020.

\bibitem[Hu et~al.(2021)Hu, Shen, Wallis, Allen-Zhu, Li, Wang, Wang, and
  Chen]{hu2021lora}
Edward~J. Hu, Yelong Shen, Phillip Wallis, Zeyuan Allen-Zhu, Yuanzhi Li, Shean
  Wang, Lu~Wang, and Weizhu Chen.
\newblock Lora: Low-rank adaptation of large language models, 2021.

\bibitem[Kingma \& Welling(2022)Kingma and Welling]{kingma2022autoencoding}
Diederik~P Kingma and Max Welling.
\newblock Auto-encoding variational bayes, 2022.

\bibitem[Kirstain et~al.(2023)Kirstain, Polyak, Singer, Matiana, Penna, and
  Levy]{kirstain2023pickapic}
Yuval Kirstain, Adam Polyak, Uriel Singer, Shahbuland Matiana, Joe Penna, and
  Omer Levy.
\newblock Pick-a-pic: An open dataset of user preferences for text-to-image
  generation, 2023.

\bibitem[Luccioni et~al.(2023)Luccioni, Akiki, Mitchell, and
  Jernite]{luccioni2023stable}
Alexandra~Sasha Luccioni, Christopher Akiki, Margaret Mitchell, and Yacine
  Jernite.
\newblock Stable bias: Analyzing societal representations in diffusion models.
\newblock \emph{arXiv preprint arXiv:2303.11408}, 2023.

\bibitem[Mokady et~al.(2022)Mokady, Hertz, Aberman, Pritch, and
  Cohen-Or]{mokady2022nulltext}
Ron Mokady, Amir Hertz, Kfir Aberman, Yael Pritch, and Daniel Cohen-Or.
\newblock Null-text inversion for editing real images using guided diffusion
  models, 2022.

\bibitem[Park et~al.(2019)Park, Liu, Wang, and Zhu]{park2019semantic}
Taesung Park, Ming-Yu Liu, Ting-Chun Wang, and Jun-Yan Zhu.
\newblock Semantic image synthesis with spatially-adaptive normalization, 2019.

\bibitem[Rabe \& Staats(2022)Rabe and Staats]{rabe2022selfattention}
Markus~N. Rabe and Charles Staats.
\newblock Self-attention does not need $o(n^2)$ memory, 2022.

\bibitem[Radford et~al.(2016)Radford, Metz, and
  Chintala]{radford2016unsupervised}
Alec Radford, Luke Metz, and Soumith Chintala.
\newblock Unsupervised representation learning with deep convolutional
  generative adversarial networks, 2016.

\bibitem[Radford et~al.(2021)Radford, Kim, Hallacy, Ramesh, Goh, Agarwal,
  Sastry, Askell, Mishkin, Clark, Krueger, and Sutskever]{radford2021learning}
Alec Radford, Jong~Wook Kim, Chris Hallacy, Aditya Ramesh, Gabriel Goh,
  Sandhini Agarwal, Girish Sastry, Amanda Askell, Pamela Mishkin, Jack Clark,
  Gretchen Krueger, and Ilya Sutskever.
\newblock Learning transferable visual models from natural language
  supervision, 2021.

\bibitem[Razavi et~al.(2019)Razavi, van~den Oord, and
  Vinyals]{razavi2019generating}
Ali Razavi, Aaron van~den Oord, and Oriol Vinyals.
\newblock Generating diverse high-fidelity images with vq-vae-2, 2019.

\bibitem[Reed et~al.(2016)Reed, Akata, Yan, Logeswaran, Schiele, and
  Lee]{reed2016generative}
Scott Reed, Zeynep Akata, Xinchen Yan, Lajanugen Logeswaran, Bernt Schiele, and
  Honglak Lee.
\newblock Generative adversarial text to image synthesis, 2016.

\bibitem[Rombach et~al.(2022)Rombach, Blattmann, Lorenz, Esser, and
  Ommer]{rombach2022highresolution}
Robin Rombach, Andreas Blattmann, Dominik Lorenz, Patrick Esser, and Björn
  Ommer.
\newblock High-resolution image synthesis with latent diffusion models, 2022.

\bibitem[Ruiz et~al.(2023)Ruiz, Li, Jampani, Pritch, Rubinstein, and
  Aberman]{ruiz2023dreambooth}
Nataniel Ruiz, Yuanzhen Li, Varun Jampani, Yael Pritch, Michael Rubinstein, and
  Kfir Aberman.
\newblock Dreambooth: Fine tuning text-to-image diffusion models for
  subject-driven generation, 2023.

\bibitem[Sohn et~al.(2023)Sohn, Ruiz, Lee, Chin, Blok, Chang, Barber, Jiang,
  Entis, Li, Hao, Essa, Rubinstein, and Krishnan]{sohn2023styledrop}
Kihyuk Sohn, Nataniel Ruiz, Kimin Lee, Daniel~Castro Chin, Irina Blok, Huiwen
  Chang, Jarred Barber, Lu~Jiang, Glenn Entis, Yuanzhen Li, Yuan Hao, Irfan
  Essa, Michael Rubinstein, and Dilip Krishnan.
\newblock Styledrop: Text-to-image generation in any style, 2023.

\bibitem[Song et~al.(2021)Song, Sohl-Dickstein, Kingma, Kumar, Ermon, and
  Poole]{song2021scorebased}
Yang Song, Jascha Sohl-Dickstein, Diederik~P. Kingma, Abhishek Kumar, Stefano
  Ermon, and Ben Poole.
\newblock Score-based generative modeling through stochastic differential
  equations, 2021.

\bibitem[Tang et~al.(2023)Tang, Rybin, and Chang]{tang2023zerothorder}
Zhiwei Tang, Dmitry Rybin, and Tsung-Hui Chang.
\newblock Zeroth-order optimization meets human feedback: Provable learning via
  ranking oracles, 2023.

\bibitem[Wu et~al.(2023)Wu, Sun, Zhu, Zhao, and Li]{hps}
Xiaoshi Wu, Keqiang Sun, Feng Zhu, Rui Zhao, and Hongsheng Li.
\newblock {Better Aligning Text-to-Image Models with Human Preference}.
\newblock \emph{ArXiv}, abs/2303.14420, 2023.

\bibitem[Xu et~al.(2023)Xu, Guo, Wang, Huang, Essa, and Shi]{xu2023promptfree}
Xingqian Xu, Jiayi Guo, Zhangyang Wang, Gao Huang, Irfan Essa, and Humphrey
  Shi.
\newblock Prompt-free diffusion: Taking "text" out of text-to-image diffusion
  models, 2023.

\bibitem[Zhang(2023)]{zhang2023major}
Lvmin Zhang.
\newblock [{Major} {Update}] {Reference}-only {Control} ·
  {Mikubill}/sd-webui-controlnet · {Discussion} \#1236, 2023.
\newblock URL
  \url{https://github.com/Mikubill/sd-webui-controlnet/discussions/1236}.

\end{thebibliography}
